\documentclass[runningheads]{llncs}

\usepackage[review,year=2024]{eccv}
\usepackage{eccvabbrv}
\usepackage{graphicx}
\usepackage{booktabs}
\usepackage[accsupp]{axessibility} 
\usepackage[pagebackref,breaklinks,colorlinks,citecolor=eccvblue]{hyperref}

\usepackage[dvipsnames]{xcolor}
\usepackage{array}
\newcommand{\STAB}[1]{\begin{tabular}{@{}c@{}}#1\end{tabular}}
\newcommand*{\ditto}{---\texttt{"}---}
\usepackage{enumitem}

\begin{document}

\title{Premonition: Using Generative Models to Preempt Future Data Changes in Continual Learning} 

\titlerunning{~}

\author{Mark D. McDonnell\inst{1} \and
Dong Gong\inst{2} \and
Ehsan Abbasnejad\inst{1} \and \\Anton van den Hengel\inst{1}}

\institute{Australian Institute for Machine Learning, The University of Adelaide \and
School of Computer Science and Engineering, University of New South Wales
\email{mark.mcdonnell@adelaide.edu.au, dong.gong@unsw.edu.au, ehsan.abbasnejad@adelaide.edu.au, anton.vandenhengel@adelaide.edu.au}}

\authorrunning{~}

\maketitle

\vspace{-0.5cm}
\begin{abstract}
Continual learning requires a model to adapt to ongoing changes in the data distribution, and often to the set of tasks to be performed.  It is rare, however, that the data and task changes are completely unpredictable. Given a description of an overarching goal or data theme, which we call a {\em realm}, humans can often guess what concepts are associated with it.  We show here that the combination of a large language model and an image generation model can similarly provide useful premonitions as to how a continual learning challenge might develop over time.  We use the large language model to generate text descriptions of semantically related classes that might potentially appear in the data stream in future.  These descriptions are then rendered using Stable Diffusion to generate new labelled image samples. The resulting synthetic dataset is employed for supervised pre-training, but is discarded prior to commencing continual learning, along with the pre-training classification head. We find that the backbone of our pre-trained networks can learn representations useful for the downstream continual learning problem, thus becoming a valuable input to any existing continual learning method. Although there are complexities arising from the domain gap between real and synthetic images, we show that pre-training models in this manner improves multiple Class Incremenal Learning (CIL) methods on fine-grained image classification benchmarks. Supporting code can be found at \url{https://github.com/cl-premonition/premonition}.
%\keywords{Continual learning \and Class Incremental Learning (CIL)}
\end{abstract}

\section{Introduction}\label{S:Intro}

In Continual Learning (CL)
~\cite{Gido22,wang2023comprehensive,buzzega2020dark,yan2022learning} the distribution of the accessible training data changes over time, and in some cases the set of tasks to be carried out changes also.  The motivation for CL is the case where it is impractical to retain the entire dataset and retrain (or finetune) periodically. However, it is known that without access to past data,  \emph{catastrophic forgetting}
~\cite{Gido22,wang2023comprehensive} tends to occur, meaning that model performance on past tasks degrades as new tasks are added.

There has been an assumption in the CL literature that the changes to the data distribution and the set of tasks to be performed are unpredictable. However, real-world task streams often appear with inherent correlations that follow some latent generic rules. As a result, it is possible to predict the kinds of challenges an agent might face in many real-world cases.  We introduce {\em realm} as a term that describes the overarching goal or data theme in a CL problem. If the {\em realm} is domestic robotics, for instance, it is likely that different types of furniture  will need to be recognised.  To ignore this fact, and wait unprepared until the data arrives, inevitably lowers practical performance. Given the power of large language models (LLMs)~\cite{Devlin2019,openai2023gpt4} to model the associations humans have between concepts, we investigate their ability to provide useful premonitions as to those that a CL agent might face. We specifically consider here the case where a CL model must learn to identify new classes of objects depicted in images, as it is one of the more challenging forms of CL. This subclass of CL is known as Class Incremental Learning (CIL) \cite{Gido22,yan2022learning}. 
\begin{figure}[tb]
  \centering
  \includegraphics[width=0.9\columnwidth]{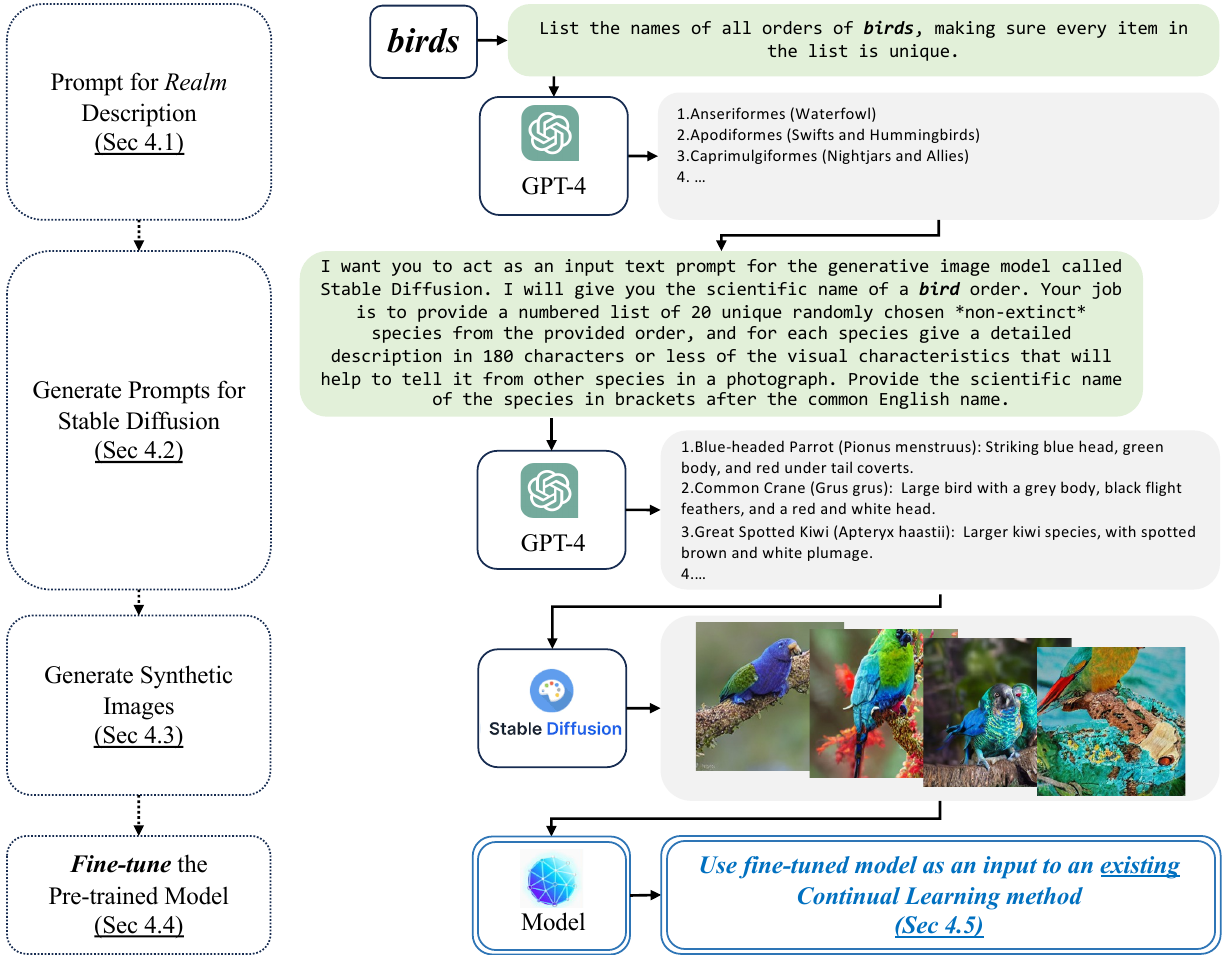}
  \caption{{\bf Overview of {\em Premonition}.} {\em Premonition} involves first prompting GPT-4~\cite{openai2023gpt4} to produce a set of text prompts containing a class name and description within a semantic {\em realm} (in this example, we use {\bf \em birds} as the realm).  It then provides these prompts as inputs to Stable Diffusion, in order to generate a dataset of synthetic imagery. {\em Premonition} then adapts a pre-trained classification model by careful transfer learning on the synthetic dataset. The resulting backbone network becomes the input to an {\em existing} CL method, ideally one designed for pre-trained models~\cite{zhou2024continual}.}\label{fig:1}
\end{figure}
 The emergence of language-conditioned image generation models \cite{DALL-E,Rombach_2022_CVPR} also makes it natural to consider whether synthetically generated data can be used in place of unavailable data from past tasks~\cite{jodelet2023classincremental}, since if the synthetic data were truly from the same distribution as the past data and subsequently used in training, then forgetting will be avoided. Most existing literature suggests this idea is unlikely to succeed, because classifiers trained on synthetic image data typically have significantly reduced performance when applied to real imagery, an effect dubbed the ``synthetic gap''~\cite{Zhang_synthetic_gap}.  However,~\cite{Sariyildiz_2023_CVPR} recently demonstrated a new and more fruitful way to exploit synthetic imagery for image classification, finding that supervised learning on a fully synthetic version of the ImageNet~\cite{ImageNet} (IN1K) dataset results in a network with very good {\em transfer learning} performance. This suggests that despite the domain gap between real and synthetic imagery, training on well-designed synthetic datasets can result in a model equipped with useful feature representations. We therefore hypothesise that CL model performance might also be improved by pre-training on appropriately-selected synthetic data.

Consistent with past CIL research, we aim to test this approach on `split' versions of common image classification datasets. However, many of the standard image classification datasets were not collected for a particular scenario-based objective. Thus, when split to task sequences in CIL, each task tends to cover a broad range of topics relatively sparsely, without an informative latent realm that is common in real-world applications. As a result, they cannot reveal the benefits of the strategy we propose here. To demonstrate the gains of our approach with conciseness, we therefore resort to experiments conducted on \emph{fine-grained} image classification datasets. Such data would be very difficult for most existing CIL methods to perform well, with the exception of those that leverage pre-trained models~\cite{zhou2024continual}, and hence these are the target for our experiments.

Based on the preceding observations, in this paper we propose {\em Premonition}, an approach to pre-training with LLM-guided synthetic data (\cref{fig:1}), and  demonstrate its value for CL on fine-grained classification datasets. 
The LLM-based guidance is used to obtain concepts possibly related to a {\em realm}, which in turn become part of prompts for image generation. Pre-training on the generated synthetic data adjusts the representation space of models that subsequently become inputs to CIL methods, resulting in improved performance. 
Note that the proposed method does not rely  on accurate prediction of unseen future tasks/classes;  the synthetic data provides informative guidance to shape a model's representation space as  preparation for future tasks. 
We report strong results despite the synthetic images generated using our LLM descriptions resulting in poor zero shot performance using the same set of language prompts, while also being easy to  discriminate from real imagery using a trained binary classifier. Overall, our contributions are as follows: 
\begin{enumerate}
\item We propose {\em Premonition}: a method for generating and exploiting synthetic pre-training datasets that leverages the knowledge embedded in a large language model and a generative image model.
\item We show that application of {\em Premonition} prior to the use of existing CIL methods for pre-trained models produces significantly better results than the vanilla versions of those methods. 
\item On some datasets, we show that even models trained from scratch on the synthetic data are able to achieve CIL performance superior to vanilla CIL methods that start with IN1K weights.
\item We also show that {\em Premonition} is vastly superior to an alternative strategy of replacing real data from past tasks  with synthetic data during CL training. 

\end{enumerate}

\section{Related work}\label{S:Related}

\subsection{EFCIL Using Pre-trained Models}\label{S:EFCIL}

Significant progress has been made recently in Exemplar-Free Class Incremental Learning (EFCIL)~\cite{Petit23_FeTriL} (or rehearsal-free CL~\cite{smith2023closer}), especially image classification based on models pre-trained on large public datasets such as ImageNet~\cite{hayes2020lifelong,Hocquet20,mehta2021empirical,Jie_2022_CVPR,Wu_2022_CVPR,panos2023session,L2P,pelosin2022simpler,Ermis22,L2P,DualPrompt,smith2023codaprompt,janson2023simple,zhou2023revisiting,zhang2023slca,smith2023closer,S-prompts,liu2023class_language,Gao_2023_ICCV,Khan_2023_ICCV,marouf2023rethinking,petit2023analysis,panos2023session,McDonnell23,zhou2024continual}. The benefit of pre-trained models, when compared to methods designed to learn continually using a model trained from scratch, is that they are good generic feature extractors, meaning that CIL methods can be designed that focus on appending relatively few parameters while the pre-trained model remains frozen either throughout CIL training, or after the first task is completed (`first session adaptation')~\cite{panos2023session,zhou2023revisiting,McDonnell23}. Amongst the cited papers, the main choice of architecture has been ViT-B/16 models and ResNet50 models. Many different pre-training choices have been used, including both self-supervised and supervised models, with different choices having a significant impact on performance~\cite{petit2023analysis}.

\subsection{Language Models for EFCIL Based on Pre-trained Models}

Several works have proposed CIL methods that leverage a combination of frozen vision and language models from CLIP~\cite{S-prompts,Zhou_CLIP,liu2023class_language} (this is distinct from work on continual training of the CLIP models as a direct goal~\cite{ding2022dont,garg2023ticclip,ni2023continual}). A recent paper built on methods that continually learn prompts for ViT models~\cite{L2P,DualPrompt} by introducing  language guidance from an arbitrarily pre-trained language model into the learning process~\cite{Khan_2023_ICCV}. In~\cite{jodelet2023classincremental}, Stable Diffusion was used to generate synthetic imagery but, different to {\em Premonition}, the data was used {\em throughout} CIL training (including as rehearsal exemplars), within existing CIL methods iCARL, FOSTER and LUCIR. 

\subsection{Transfer Learning for CIL}

Our approach, {\em Premonition}, resembles past use of self-supervised learning to pre-train a model prior to transfer to CIL~\cite{gallardo2021self,liu2023class}. However~\cite{gallardo2021self} assumed the data available for self-supervision is a subset of the data subsequently used in training the CIL model, and~\cite{liu2023class} assumes that the first CIL task is much bigger than subsequent CIL tasks. Differently, we tackle the more difficult problem where the entirety of benchmark CIL datasets are split into relatively small subsets as independent training tasks, while in our pre-training on synthetic data, we assume \underline{no} knowledge of any downstream CIL class names. Other CIL approaches use a combination of self-supervised and supervised learning within each task~\cite{Cha_2021_ICCV,ni2023continual,pham2021dualnet,bhat2022task} or continually update a pre-trained model during CIL training, requiring multiple methods to combat forgetting, including non-EFCIL use of rehearsal memory buffers~\cite{Boschini}.

\subsection{Training Models with Synthetic Data}

The emergence of powerful language-guided generative models such as DALL-E~\cite{DALL-E} and Stable Diffusion~\cite{Rombach_2022_CVPR} has sparked renewed enthusiasm for using photo-realistic synthetic imagery to  train other models such as classifiers~\cite{shipard2023diversity,Sariyildiz_2023_CVPR,bansal2023leaving,azizi2023synthetic,trabucco2023effective,dunlap2023diversify}.  The appeal of doing so is clear:  having complete control over the process that generates the training data for a model would address many of the limitations of real training sets, such as the cost of labelling. Synthetic training data might represent the classes of interest in more detail, or even depict classes that are impractical to capture in real data, for example.  However, the mentioned synthetic gap~\cite{Zhang_synthetic_gap} remains stark. Subtle differences between synthetic and real images may be imperceptible to humans while negatively impacting model performance~\cite{bird2023cifake}, and making synthetic imagery easy to detect~\cite{wu2023generalizable,lorenz2023detecting,rahman2023artifact}. Recent investigations using Stable Diffusion~\cite{Rombach_2022_CVPR} have reported that although the gap is closing, particularly when synthetic data augments real data~\cite{bansal2023leaving,azizi2023synthetic,trabucco2023effective,dunlap2023diversify},  it remains problematic~\cite{shipard2023diversity,Sariyildiz_2023_CVPR}. In one investigation~\cite{shipard2023diversity},  synthetic CIFAR-10 and CIFAR-100 datasets were created from Stable Diffusion using prompts based on the CIFAR class names, and  used to train a model tested on the corresponding real test sets. This approach was characterized as zero-shot classification, since no actual real CIFAR data is used in training. The test-set error rate compared to supervised training on the real datasets is in the order of five times larger. Similarly, in~\cite{Sariyildiz_2023_CVPR}, synthetic ImageNet clones are created from Stable Diffusion and used to train classifiers for the real ImageNet test set. Again, the error rates on the real test set are 2-3 times higher than the baseline. However, as mentioned,~\cite{Sariyildiz_2023_CVPR} also showed that the synthetic ImageNet clones had clear strong value when applied to transfer learning, a finding which is key to {\em Premonition}.

Synthetic data  used for EFCIL in one investigation~\cite{Smith_2021_ICCV} differs from {\em Premonition} in two ways: first, the synthetic data was extracted by an inversion method from the CIL classifier, and therefore reliant on the CIL training set. Second, synthetic data was used throughout CIL training.

\subsection{CIL Using Data Replacement}

In Section~\ref{S:Strategies}, we describe two strategies for the use of Synthetic Data in CIL, one of which we call Data Replacement. We already mentioned an approach that does this~\cite{jodelet2023classincremental}.  Other Data Replacement strategies do not use synthetic data. First, a common CIL approach is to use rehearsal/replay methods that assume the possibility of selecting and storing exemplars from the training sets of past tasks~\cite{Gido22,wang2023comprehensive}. As mentioned, we assume EFCIL, \ie that this is not possible. Differently, a recent proposal is to use uncurated `webly-supervised' data. This can be considered an extreme form of Data Replacement, whereby no curated data at all is assumed to exist for CIL training. It takes the form of  scraping the web for training samples at the point in time when a list of the names of classes for new tasks becomes known~\cite{prabhu2023categories}; this approach is called Continual Name-Only Classification. {\em Premonition} assumes the classical CIL scenario in which sets of curated  CIL training data become available with each new set of class names, and hence that there is no need to use Name-Only Classification.
\section{CL Scenario and Assumptions}

In this section we list the attributes and assumptions of the CL problem scenario tackled in this paper. Specifics of our {\em Premonition} methods are given in~\cref{S:Methods}.

\subsection{A Known {\em Realm} for a Continual Learning Agent}

{\em Premonition} uses an LLM as a knowledge base that lists potentially relevant classes from the task context, expressed as a short natural language description of an overarching goal or data theme relevent to a CL agent (we call this a {\em realm}), and then generates synthetic images in these classes. More formally, we define {\em realm} as the abstraction of the context of the problem where class labels (and consequently images) are conditionally generated. 

\subsection{Premonition Does Not Know Future Classes}

We emphasise that it is not known in advance how many classes and what class names will be needed during CIL. Hence the CIL model will not be trained on the same set of classes as those sampled conditioned on a {\em realm}. Nevertheless, we expect reasonable overlap due to the universality of LLMs. As an example, if the {\em realm} is ``birds'', our model will be pre-trained to discriminate synthetic data from different bird species. But these synthetic images are {\em only used for supervised pre-training} and then discarded subsequently, along with the pre-training classification head. Only the resulting pre-trained backbone model is retained, and this becomes an input to existing CIL methods.  This is why our synthetic data has no need to overlap the same classes as the real CIL data.

\subsection{Exemplar-Free Class Incremental Learning (EFCIL)}

We choose to investigate pre-training on synthetic data for CIL~\cite{zhou2023class} as, of the three primary variants of CL, this assumes the least of the environment (it does not assume task labels are available, for example)~\cite{Gido22}. 
We also assume EFCIL, \ie where training does not use rehearsal/replay of exemplars from past tasks.

\subsection{Pre-training Instead of Data Replacement}\label{S:Strategies}

We consider two ways in which synthetic data could be used for CIL:
\begin{description}[align=left,leftmargin=*,labelsep=\parindent]
\item[Strategy 1 -- Data replacement:] At the start of training for each CL task, generate new synthetic data to replace unavailable data from old tasks.
\vspace{0.2cm}
\item[Strategy 2 -- {\em Premonition}:] Generate a synthetic dataset {\em before} commencing CL training; use this to train a model using supervised learning;  discard the classifier head from pre-training, and use the backbone as a pre-trained network. 
\end{description}
In~\cref{S:Results}, we present evidence that despite not having prior knowledge of specific classes encountered in CIL training, {\em Premonition} significantly outperforms Strategy 1, which does have that knowledge, due to the domain gap between real and synthetic imagery. Moreover  Strategy 1 requires significantly more compute when updating models, which is untenable for streaming data scenarios~\cite{Prabhu23}.% We discuss the details of our implementation of {\em Premonition} in~\cref{S:Methods}.

\subsection{No First-Session Adaptation}

We assume a more difficult CL scenario than in some related work. Specifically, we rule-out the possibility of using `first-session' adaptation, as studied by~\cite{panos2023session,zhou2023revisiting,McDonnell23}. The rationale for this is that a first task needs to be of sufficiently large size for the technique to be effective, and there is no guarantee of this in practice. It also is untenable for streaming data~\cite{Prabhu23}. Under the assumption that first-session adaptation cannot be used, we aim to show that `priming' a model on synthetic data can replace at least some of the benefits of `first-session'  adaptation of a pre-trained model to a downstream CIL dataset.

\subsection{Fine-grained Image Classification}

We already discussed in the Introduction why {\em fine-grained}~\cite{fine_grained} image classification datasets form a good test for {\em Premonition}. Several additional reasons follow. 
First, preliminary work on split CIFAR-100 indicated that pre-training on synthetic versions of small diverse-concept datasets like CIFAR-100 do not transfer well, while better performance of synthetic ImageNet datasets can require considerably more samples than the real dataset~\cite{Sariyildiz_2023_CVPR} (see~\cref{S:split}). We hypothesised that a way to produce a stronger model with a relatively small synthetic dataset was to pre-train on a relatively large number of semantically similar classes, as in fine-grained datasets. Second,  we use Stable Diffusion for data generation, and this model relies on the CLIP vision-language model~\cite{Radford}. CLIP is trained on a large dataset of 400M samples, covering a broad range of visual concepts~\cite{Radford}, and Stable Diffusion on over 1B images. Datasets of common concepts like in CIFAR-100 and ImageNet are more likely to have many of their training samples well represented within the training data of CLIP and Stable Diffusion. When coupled with the fact that methods exist for extracting Stable Diffusion's training images from its trained models~\cite{carlini2023extracting},  creating synthetic CIFAR or ImageNet datasets runs a higher risk than fine-grained synthetic datasets of creating train-validation contamination, and so is best avoided when attempting to draw general conclusions. Finally,  CLIP is well-known to be a very strong zero-shot classifier~\cite{Radford}. Since we use its vision model in Stable Diffusion, it is reasonable to ask why not simply use CLIP in zero-shot mode as a classifier instead of training a CIL model? This might be valid for datasets like CIFAR-100, but CLIP zero-shot classification has been found to perform relatively poorly on fine-grained classification tasks compared with more general datasets~\cite{Radford}. 

\section{Premonition}\label{S:Methods}

The results of~\cite{Sariyildiz_2023_CVPR} on the transferability of synthetic ImageNet clones lead us to a CIL strategy called {\em Premonition} (\cref{fig:1}). The core idea is to  {\em pre-train on synthetic data} before applying existing CIL methods for pre-trained models.

\subsection{Generation of Text Prompts From a LLM}\label{S:GPT4}

 In the first step, a  large language model (LLM) is used to create text prompts for later use in a generative image model. We used GPT-4~\cite{openai2023gpt4}
and  Stable Diffusion~\cite{Rombach_2022_CVPR}. Our prompting strategy is similar to previous work for which the downstream task was CLIP zero-shot classification~\cite{menon2022visual}.  To begin, assuming only the {\em realm} is known in advance, we  prompt GPT-4 with a `user prompt' to generate a large set of subtypes within the {\em realm}. For example, for the {\em realm} `birds' we desire broad coverage of all orders (sub-classes) of birds, and use: 
\begin{quote}
{\tt List the latin names of all orders of birds, making sure every item in the list is unique.}
\end{quote}
For each returned bird family, we then prompt GPT-4 with a `system' prompt of the form: \begin{quote}
    {\tt I want you to act as an input text prompt for the generative image model called Stable Diffusion. I will give you the \\scientific name of a bird order. Your job is to provide a \\numbered list of 20 unique randomly chosen *non-extinct* \\ species from the provided order, and for each species give a detailed description in 180 characters or less of the visual characteristics that will help to tell it from other species in a photograph.}
\end{quote}
The full prompt used has some further instructions to help ensure desired formatting and extraneous remarks are omitted from the response---see~\cref{SM:prompts}. The `system' prompt is then combined with a `user' prompt consisting of a single bird order name to trigger lists of generated species descriptions. For some comparison experiments, we assume  class names are known in advance. We use a slightly different `system' prompt for this---see~\cref{SM:prompts}.   

\subsection{Text Prompts for Text-to-Image Generation}\label{S:text_prompts}

In most experiments, our prompts are of the form: {\tt A photograph of a *class name*: *class description*}. In some experiments, we omit the description and use: {\tt A photograph of a type of *{\em realm} name*: *class name*}.

\subsection{Generation of Images from Stable Diffusion}

For text-to-image generation, we used Stable Diffusion~\cite{Rombach_2022_CVPR}. As our use case for the images was to train a classifier, we did not require large high resolution images to be produced, instead seeking a typical classifier input size of 256 pixels. To this end, we used a pre-existing (frozen) `mini' version of Stable Diffusion 1.4~\footnote{\tt huggingface.co/lambdalabs/miniSD-diffusers} that by default generates images of size $256\times256$ pixels, working with a latent of size $32\times 32\times4$. This contrasts with standard versions of Stable Diffusion for which the default output size is  $ 512\times512$ pixels, and the latent spatial dimension is $64$.  This choice enabled us to generate synthetic imagery much faster, and reduce the storage overhead and preprocessing overhead times for downstream models.  Like the standard Stable Diffusion 1.4 model, the language encoder is the frozen CLIP ViT-L/14 model, with embedding size 768. We set the number of Stable Diffusion inference steps to 40. As recommended by~\cite{Sariyildiz_2023_CVPR}, we use a guidance scale of 2.0. For each distinct input prompt, we generate multiple examples using identical Stable Diffusion inputs, and rely on the randomness of the model to generate data diversity.

\subsection{Pre-training on Synthetic Datasets}\label{S:resnet}

Our experiments focus  on  the ResNet50 architecture~\cite{he2015deep}, due to the prevalence of available pre-trained models. We compared models (i) pre-trained using supervised learning on ImageNet1K (IN1K); (ii) pre-trained using language-guided self-supervision on RedCaps (LGSSL)~\cite{El_Banani_2023_CVPR}; (iii) pre-trained purely on synthetic ImageNet data (FITYMI)~\cite{Sariyildiz_2023_CVPR}; and (iv) pre-trained purely on our own synthetically generated datasets from scratch. %For ViT-B/16 we used a model pre-trained on ImageNet21K using self-supervised learning (IN21K).
 Note that this {\em Premonition} step is not CIL, so we trained jointly on all classes in the synthetic dataset using commonly adopted parameter choices (see~\cref{SM:PT}).

\subsection{Class-Prototype Methods for CIL Based on Pre-trained Models}\label{S:CIL_outline}
After {\em Premonition}'s supervised pre-training on synthetic data is completed, the resulting model (minus the discarded classification head) becomes an input to the standard CIL scenario. This assumes a sequence of $T$ tasks, $\mathcal{D}=\{\mathcal{D}_1,\dots\mathcal{D}_T\}$, so that each $\mathcal{D}_t$ is a disjoint set of labelled training samples to  be learned from. Training for each task can only access its own training data. During inference, the task ID is unknown, but the model is tested as if jointly trained on all tasks.

For testing {\em Premonition}, we use existing CIL methods for pre-trained models~\cite{zhou2024continual}: Nearest Class Mean (NCM)~\cite{janson2023simple,panos2023session,zhou2023revisiting,McDonnell23}, Continual LDA~\cite{panos2023session} and RanPAC~\cite{McDonnell23}. These three methods are class-prototype based (see~\cref{SM:CIL} for an overview); classification is done in a manner that avoids forgetting, by extracting feature representations from the model, finding the mean representation in each class, and selecting the class for which this representation has highest similarity with the representation for test samples~\cite{McDonnell23}. Although~\cite{panos2023session,McDonnell23} use first session adaptation, as discussed above we omit that component of those methods.  For RanPAC, we modified the method of~\cite{McDonnell23} to improve its performance on datasets with severe class imbalance;  details are in~\cref{SM:CIL}.

\section{Results}\label{S:Results}

\subsection{Prompts Created by GPT-4}\label{S:prompt}

An example `birds' prompt generated by GPT-4 is
\begin{quote}
{\tt Pionus menstruus (Blue-headed Parrot):  Medium-sized parrot \\with blue head, green body, and red undertail coverts.}
\end{quote} 

\subsection{Synthetic Datasets from Stable Diffusion}

\begin{figure}[tb]
  \centering
{\includegraphics[width=0.25\columnwidth,height=0.25\columnwidth]{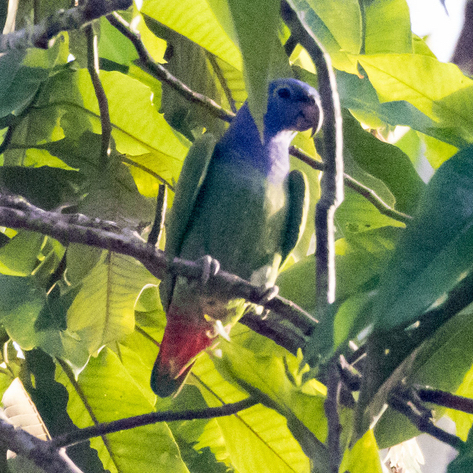}
\hspace{-1mm}\includegraphics[width=0.25\columnwidth,height=0.25\columnwidth]{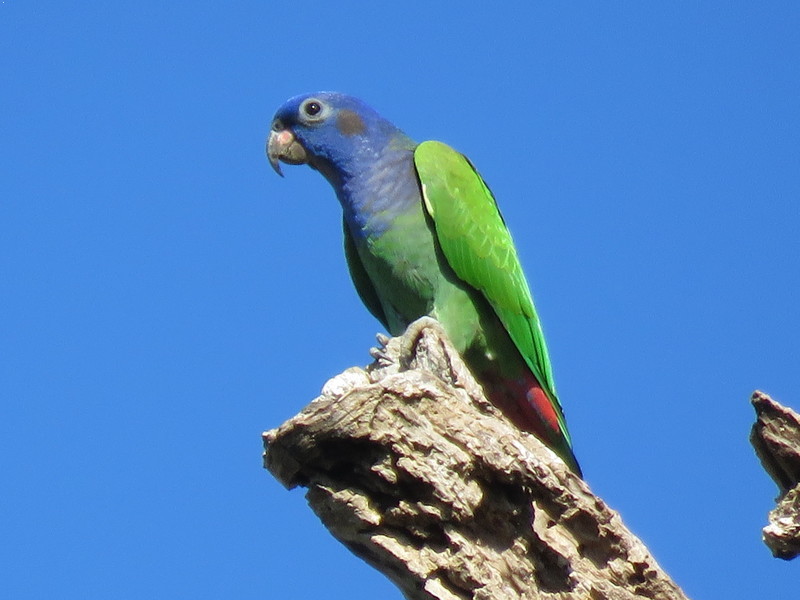}}\\
   %\vspace{-1mm}
   {
  \includegraphics[width=0.25\columnwidth]{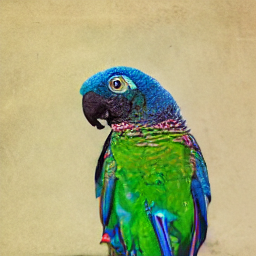}\hspace{-1mm}
   \includegraphics[width=0.25\columnwidth]{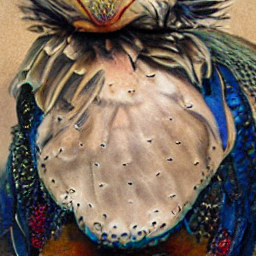}}
  \caption{{\bf Comparison of real images with synthetic images.} The two images in the top row are for the species Pionus menstruus, obtained from iNaturalist-2018~\cite{inat}. The two images in the bottom row are examples generated by Stable Diffusion from the prompt example given in~\cref{S:prompt}. The second synthetic image is clearly a failure case, and many others were evident. Nevertheless, we found that {\em Premonition}  produces benefits despite being trained on images that sometimes do not resemble the real concepts as well as they ideally should.}\label{fig:SD}
\end{figure}

For convenience, we created and stored on disk three primary synthetic datasets corresponding to three {\em realms}: `birds', `food' and `plants'. Details are in~\cref{SM:synthetic_datasets}, and examples of two generated images are in~\cref{fig:SD}, compared with two real images from the same bird species. In our experiments, we  trained on multiple epochs of these datasets; a valid alternative would be to synthesize an endless sequence of new images during training.

\subsection{Premonition for Fine-Grained  Classification }

The primary results of this paper are shown in~\cref{T_primary}.  We test the value of {\em Premonition} when added to three different class-prototype methods designed for CIL based on pre-trained models, as outlined in~\cref{S:CIL_outline}, and for multiple choices for pre-trained weights (inputs to {\em Premonition}).  The metric reported is Final Average Accuracy~\cite{NIPS2017_f8752278}: given classification accuracies, $R_{t,i}$, on the $i$--th task,  following training on the $t$-th task, Average Accuracy is defined as  $A_t=\frac{1}{t}\sum_{i=1}^tR_{t,i}$ and Final Average Accuracy as $A_T$.
 
Recall that {\em Premonition} does not assume specific future classes can be anticipated, and knows only the {\em realm} of the classes, namely `birds', `food' and `plants.'  We create a single synthetic dataset for each realm, and then test CIL performance on at least three fine-grained real datasets within each {\em realm}, with each realm having at least one dataset with over 1000 classes (detailed in~\cref{SM:real_datasets}). For `birds', we use CUB~\cite{CUB}, NABird~\cite{NAbird}, the birds subset from Omnibenchmark (OB)~\cite{OmniBenchmark} (OB-bird) and the aves subset of iNaturalist-2018~\cite{inat}  (iNat-2018-aves). For `food', we use Food-101~\cite{bossard14}, OB-food, and Food2K~\cite{food2k}. For `plants', we use 102 Flowers~\cite{102flowers},  OB-plants and Pl@ntNet-300K~\cite{plantnet-300k}. % and  the subset of iNaturalist (2018) for plantae (iNat-plantae). 
For each dataset, we used the standard train/test splits, and for each we created CIL variants, by randomly assigning an equal number of classes to each of $T=10$ tasks. Any leftover classes were assigned to the first task. 

Our primary findings in~\cref{T_primary} are  that for the cases where {\em Premonition} starts with IN1K and FITYMI weights, it improves nearly every dataset/CIL method, often by large margins. In particular, for RanPAC~\cite{McDonnell23} (the strongest of the three CIL methods), the raw accuracy boost for IN1K is  between $4.6-7.9$\% for all datasets in the `birds' {\em realm}, and between $3.7-7.0$\% for all datasets in the `plants' {\em realm}. For `food', a gain of $2.3\%$ for Food-101 and $7.3$\% is seen for Food2K. Supplementary data in~\cref{SM:cross} shows that it is important that {\em Premonition} uses synthetic data from its correct realm, or otherwise accuracy tends to be worse than the baseline models where {\em Premonition} is not used. Conversely, if instead  the synthetic data is an aggregation of all three of our example {\em realms}, performance is actually enhanced on the downstream CIL datasets compared with using a single {\em realm} for pre-training. Hence, {\em Premonition} should be capable of application to CL agents that learn from multiple {\em realms}.

For the third pre-trained model, LGSSL, the inconsistency of the results in~\cref{T_primary} is enlightening. For the `food' and `plants' {\em realms}, although the use of {\em Premonition} causes accuracy to go backwards, the baseline LGSSL model is stronger than the other two, while when using {\em Premonition} the model initialized with LGSSL is still better than the other initializations for `food.'  A likely explanation for this is that RedCaps is a much larger dataset than IN1K, and includes nearly 600K training samples for both food and plant related concepts. This suggests the model is already well primed for food and plants datasets, and why {\em Premonition} does not add value.  On the other hand, {\em Premonition} does provide strong improvements to LGSSL for all `birds' datasets. We conclude that {\em Premonition} is more likely to have value when pre-trained models are not already very well specialized for a particular theme.

{\em Premonition} is primarily designed to enhance existing pre-trained models, but as shown in the {\bf Scratch training} section of~\cref{T_primary}, if a model is instead trained from scratch, we can still achieve good CIL outcomes. Interestingly, in several cases, the scratch results are stronger than at least one of the baseline (non-{\em Premonition}) results for the same CIL method.  On the other hand, {\em Premonition} with transfer learning from IN1K, FITYMI or LGSSL always clearly outperforms the scratch models.

Although our experiments focus on a set of CIL methods that specialize in using representations extracted from frozen pre-trained models, {\em Premonition} also has value for other more traditional CIL methods, such as EWC~\cite{EWC} and fine-tuning~\cite{zhang2023slca}. Representative results showing {\em Premonition}-enhanced results for these methods are contained in~\cref{SM:CIL_extra}.

\begin{table}[tb]

\caption{{\bf {\em Premonition} results for three CIL methods applied to 10 datasets.} Data in the table is for Final Average Accuracies for $T=10$ CIL tasks using ResNet50.  {\bf Bold} numbers indicate cases where the use of {\em Premonition} increases accuracy relative to not using it, for the same initial pre-trained weights and CIL method. For {\em Premonition}, the listed initialization weights were updated using finetuning on the synthetic dataset corresponding to each {\em realm}. The continual learning methods applied following  {\em Premonition}  are our own re-implementations of the cited methods, as per~\cref{S:CIL_outline}.  IN1K, FITYMI, and LGSSL are defined in~\cref{S:resnet}. For {scratch training}, {\em Premonition} models were initiated with random weights instead of a generic pre-trained model.}\label{T_primary}
\centering
\begin{tabular}{p{2.4cm}>{\centering\arraybackslash}p{1.5cm}|p{0.7cm}p{0.7cm}p{0.7cm}p{0.7cm}|p{0.7cm}p{0.7cm}p{0.7cm}|p{0.7cm}p{0.7cm}p{0.7cm}}

&&\multicolumn{4}{|c|}{\bf Birds {\em realm}}& \multicolumn{3}{|c|}{\bf Food {\em realm}}& \multicolumn{3}{|c}{\bf Plants {\em realm}}\\
&
  & \STAB{\rotatebox[origin=c]{90}{CUB}} & \STAB{\rotatebox[origin=c]{90}{NABird}} & \STAB{\rotatebox[origin=c]{90}{OB-bird}} & \STAB{\rotatebox[origin=c]{90}{iNat-2018-aves}} & \STAB{\rotatebox[origin=c]{90}{Food-101}} & \STAB{\rotatebox[origin=c]{90}{OB-food}} & \STAB{\rotatebox[origin=c]{90}{Food2K}} & \STAB{\rotatebox[origin=c]{90}{~102 Flowers}} & \STAB{\rotatebox[origin=c]{90}{OB-plants}} & \STAB{\rotatebox[origin=c]{90}{~Pl@ntNet-300K}} \\
& \#  classes: & $~200$ & $~555$ & $~646$ & $1258$ & $~101$ & $~673$ & $2000$ & $~102$ & $~671$ & $1081$\\% & $2917$\\
  \bottomrule
  \end{tabular}
\vspace{-2mm}
\begin{flushleft}
{\bf Transfer learning: } 
\end{flushleft}
\vspace{-2mm}
\begin{tabular}{p{2.6cm}>
{\centering\arraybackslash}p{1.3cm}|p{0.7cm}p{0.7cm}p{0.7cm}p{0.7cm}|p{0.7cm}p{0.7cm}p{0.7cm}|p{0.7cm}p{0.7cm}p{0.7cm}}
\toprule
NCM~\cite{pelosin2022simpler,zhou2023revisiting}&IN1K&$54.1$ &$41.7$ & $28.8$& $24.9$& $52.5$&$15.4$ & $34.2$&$69.1$ &$20.3$ &$21.0$  \\
 \ditto~~ &FITYMI& $35.8$& $21.5$&$17.1$ & $11.8$& $45.9$& $12.5$& $33.7$ & $85.9$&$18.0$ & $18.7$ \\
    \ditto~~ &LGSSL&$44.6$ & $29.3$&$20.5$ &$16.2$ &$71.0$ & $19.4$&$47.3$ & $87.2$& $26.3$& $30.9$ \\
 \ditto~~+ {\bf Prem.}  &IN1K&${\bf 66.0}$ &${\bf 54.4}$ & ${\bf 34.3}$&${\bf 31.1}$ & ${\bf 57.7}$& ${\bf 15.9}$ &${\bf 45.1}$ & ${\bf 78.1}$ &${\bf 27.9}$& ${\bf 36.6}$\\
\ditto~~+ {\bf Prem.}  &FITYMI&${\bf 57.9}$ &${\bf 44.4}$ & ${\bf 27.9}$& ${\bf 22.4}$&${\bf 54.6}$ & ${\bf 14.2}$& ${\bf 43.7}$& ${\bf 88.2}$ &${\bf 27.5}$ &${\bf 35.4}$ \\
 \ditto~~+ {\bf Prem.}  &LGSSL&${\bf 56.6}$ &${\bf 41.2}$ & ${\bf 26.0}$&${\bf 20.6}$ & $57.8$& $15.9$& $40.6$&$78.6$ &$23.9$ &$30.2$  \\
  \midrule
C. LDA~\cite{panos2023session}&IN1K& $62.1$& $45.9$& $32.1$& $26.0$&$59.4$& $14.7$&$29.3$ &$81.8$ & $25.9$&$15.8$  \\
  \ditto~~ &FITYMI&$61.2$ &$47.3$ &$33.9$ & $23.9$& $64.5$&$13.7$ & $37.1$& $90.0$&$31.1$ &$21.9$ \\
     \ditto~  &LGSSL& $57.2$&$41.9$ &$32.0$ & $22.4$& $77.1$& $19.3$&$44.9$ & $86.5$& $32.9$& $25.6$ \\
  \ditto~~+ {\bf Prem.}  &IN1K&${\bf 66.9}$ & ${\bf 56.1}$& ${\bf 38.2}$& ${\bf 32.5}$& ${\bf 67.8}$& ${\bf 15.5}$ &${\bf 42.2 }$&${\bf 87.7}$  &${\bf 31.2}$ &${\bf 25.7}$ \\
\ditto~~+ {\bf Prem.}  &FITYMI&${\bf 65.3}$&${\bf 51.4}$ & ${\bf 36.1}$&${\bf 26.4}$ &${\bf 68.4}$ &  ${\bf 16.1}$&${\bf 43.7}$ & $89.6$ &${\bf 33.6}$ &${\bf 26.7}$ \\
\ditto~~+ {\bf Prem.}  &LGSSL&${\bf 61.8}$ & ${\bf 46.8}$& ${\bf 32.7}$&${\bf 23.8}$ &$70.0$ &$16.2$ &$42.0$ &${\bf 88.2}$ & $30.6$&$24.0$  \\
   \midrule
RanPAC~\cite{McDonnell23} &IN1K& $68.3$&$57.0$ & $44.8$&$37.4$ &$69.4$ &$19.5$ &$46.4$ & $83.2$& $33.0$&$37.6$  \\
 \ditto~~ &FITYMI& $60.6$& $47.8$& $42.8$&$28.8$ &$71.2$ & $18.8$&$50.7$ &$90.8$ &$39.3$ &$41.0$ \\
  \ditto~  &LGSSL&$60.0$ & $44.9$& $39.4$& $25.5$& $80.9$& $23.4$&$56.5$ &$92.6$ &$37.7$ & $47.4$ \\
%{\bf Classes} RanPAC &IN1K&79.1& & & & & & & & & & \\
\ditto~~+ {\bf Prem.}  &IN1K& ${\bf 74.7}$& ${\bf 64.9}$& ${\bf 49.4}$& ${\bf 42.2}$&${\bf 71.7}$ & ${\bf 19.5}$&${\bf 53.7}$ & ${\bf 90.2}$& ${\bf 36.7}$& ${\bf 44.3}$ \\
\ditto~~+ {\bf Prem.}  &FITYMI&${\bf 68.5}$ &${\bf 56.0}$ & ${\bf 43.4}$& ${\bf 32.4}$&${\bf 72.6}$ & $18.3$&${\bf 53.7}$ &  ${\bf 92.6 }$&${\bf 40.0}$ &${\bf 46.7}$ \\
\ditto~~+ {\bf Prem.}  &LGSSL&${\bf 64.1}$ &${\bf 51.5}$ & ${\bf 40.5}$& ${\bf 28.6}$&$73.8$ &$20.0$ &$53.2$ &$89.1$ &$35.6$ & $45.5$ \\
\bottomrule
%\vspace{3pt}
\end{tabular}
\vspace{-2mm}
\begin{flushleft}

%{\bf ResNet50 Ablations: } 
{\bf Scratch training: } 
\end{flushleft}
\vspace{-2mm}
\begin{tabular}{p{3.8cm}>{\centering\arraybackslash}p{0.1cm}|p{0.7cm}p{0.7cm}p{0.7cm}p{0.7cm}|p{0.7cm}p{0.7cm}p{0.7cm}|p{0.7cm}p{0.7cm}p{0.7cm}}
\toprule
NCM + {\bf Premonition}  && $47.1$& $31.2$& $18.7$& $12.0$ & $50.7$&$11.1$ &$34.9$ & ${82.9}$&${23.4}$ & ${26.6}$\\
C. LDA + {\bf Premonition} && $59.7$& $47.1$& $32.8$& $19.8$ & $65.6$& $12.1$&$41.0$ & ${87.1}$&${29.8}$ &${23.9}$ \\
RanPAC + {\bf Premonition} && $60.8$& $49.1$& $39.0$&$22.2$  &$70.3$ &$16.0$ & $48.6$& ${87.4}$& ${35.1}$& ${39.0}$\\
\bottomrule
\end{tabular}
\end{table}
\clearpage

\subsection{Comparison with Data Replacement for CIL}\label{S:DR}

We now compare {\em Premonition's} strategy of pre-training on synthetic data with the alternative (see~\cref{S:Strategies}) strategy of using synthetic data as a form of {\em Data Replacement}, \ie generating synthetic images that match the classes of past tasks to replace past real data within CIL. Details of the synthetic datasets created for this comparison are found in~\cref{SM:synthetic_datasets_small}. For training, we use $T=10$ tasks and  standard supervised transfer learning in which a IN1K pre-trained ResNet50 was finetuned throughout all CIL tasks using SGD. At the start of each task, we reset the optimizer and its learning rate to $0.1$ for a fully-connected output head (initialized with all $K$ classes expected after $T$ tasks) and $0.01$ for the backbone. We train for up to $15$ epochs in each task. For task $t$, the only real data are the classes corresponding to the CIL task, but our synthetic data for all classes seen in tasks $1,\dots,t$ are also trained on; thus the training set grows with each task. 

As shown in~\cref{T_dr}, we found that data replacement achieves very poor results. For example, for CUB and $T=10$ CIL tasks, only $39.1\%$ Final Average Accuracy was achieved, which is even surpassed by a large margin by our ResNet50 {\em Premonition} models trained from scratch (up to $60.8$\%). As a comparison, we also show in~\cref{T_dr} results from naive fine-tuning on each task, where no steps are taken to combat forgetting ({None}). These results are very poor, which indicates that the synthetic replacement data has some value for CIL in terms of reduced forgetting.  As a check to ensure that it is the data replacement that causes poor {Synthetic} performance in~\cref{T_dr} rather than deficiencies in the training design, we repeated the same experiments but instead of replacing past real  data with synthetic data in task $t$ we retained the real data from tasks $1,\dots,t$ and do not use synthetic data. As indicated by the strong results for {Real} in~\cref{T_dr}, the poor performance seen for {Synthetic} is attributable to using synthetic training data to replace real data. By the final task, the Real model is learning jointly from the entire real training set, and hence performs  more strongly than {\em Premonition}. 

\begin{table}[b]
\caption{{\bf CIL using data replacement.}   {\bf Premonition} denotes baseline accuracy taken  from~\cref{T_primary}.  {\bf None} denotes naive fine-tuning. {\bf Synthetic} indicates Data Replacement using synthetic data to replace past real data (from the same classes). {\bf Real} denotes the case where all past {\bf Real} data is available for joint training.}\label{T_dr}
\centering
\begin{tabular}{cc|c|c|c}
 \toprule
{\bf Dataset} &{\bf Premonition}  & {\bf None} &{\bf Synthetic} & {\bf  Real}\\
\midrule
CUB  &$74.7$& $9.0$& $39.1$&$81.4$\\
Food-101  &$68.6$&$10.3$& $20.2$&$87.6$\\
102 Flowers  &$88.0$ &$15.0$ &$46.5$ &$91.3$\\
 \bottomrule
\end{tabular}
\end{table}

As clues to why Data Replacement performs very poorly compared with {\em Premonition}, supplementary data in~\cref{SM:CLIP_ZS} shows that our synthetic data is much more difficult to  classify using CLIP zero-shot classification than real data with identical sets of class labels, which indicates the quality of the imagery falls short of the ideal. Moreover, consistent with other work~\cite{wu2023generalizable,lorenz2023detecting,rahman2023artifact}, we also find it easy to train a binary classifier to discriminate between our class-name synthetic datasets, and the corresponding real datasets used in~\cref{T_primary}; as shown in~\cref{SM:binary}, each classifier achieves very high accuracy, confirming the large domain gap between real and synthetic data.

\section{Conclusion}\label{S:Discussion}

We have shown that although our synthetic data has limited value as a data replacement strategy during CIL, their usage in supervised pre-training can endow a model with relevant feature representations highly suited to targeted transfer learning, and hence enhance continual learning methods that use these models as inputs. Overall, we hope that our proposed method -- {\em Premonition} -- and the experimental results we have demonstrated will kick-start a series of investigations into enhanced synthetic datasets with general utililty for continual learning.

\subsection{Future Work and Limitations}

Our experiments focused on demonstrating that {\em Premonition} provides a measurable increase in CIL Final Average Accuracy on multiple fine-grained datasets. There remains many possible investigations into optimizing the synthetic dataset and ultimately the pre-trained model produced by {\em Premonition}. It will be interesting for future work to consider whether new generative text-to-image models can amplify {\em Premonition}'s performance, either by better realization of the details of the text prompt within the image, or reduction of the synthetic gap.

We assume in our main experiments that {\em Premonition} has prior knowledge only of the {\em realm}. We were interested to consider whether prior knowledge of future class names would significantly boost performance. Hence, we used {\em Premonition} to generate and train on synthetic imagery for all classes in the smallest dataset in each of the three {\em realms}. The findings in~\cref{SM:class_name} suggest that advance knowledge of classes can provide a benefit over knowing only the {\em realm}. Further examination of how much more improvement could be gained is beyond our  scope. Converse to the stated assumption, a limitation of {\em Premonition} is that a clear statement of a {\em realm} may not be possible in some circumstances 

One reason for choosing to use ResNet50 models was to understand how well synthetic data could be used to pre-train a model from scratch, and compare this with the case where the model is initialized to general-purpose pre-trained models like those trained on ImageNet.  This comparison would not be as easily possible on ViT-B/16 transformer networks~\cite{vit}, used as the primary pre-trained model in much related work~\cite{L2P,zhou2023revisiting,zhang2023slca,smith2023codaprompt,McDonnell23}, as ViT-B/16 models are known to be much more difficult than ResNets to train adequately from scratch on relatively small and non-diverse datasets. However, we also conducted an exploration into applying {\em Premonition} to pre-trained ViT transformer networks. The gains we observed from using {\em Premonition} were modest, consistent with ViT pre-trained networks already being very strong generic feature extractors. Nevertheless, we suggest future work aims to identify particular {\em realms} for which ViT models are less strong and show that {\em Premonition} provides strong benefits.

\bibliographystyle{splncs04}
\bibliography{Premonition}
\clearpage
\appendix
\section{Datasets}\label{SM:datasets}

We define $K$ as the total number of classes after $T$ tasks. We define the number of training samples for task $t$ as $N_t$. Overall, the number of training samples is $N:=\sum_{t=1}^TN_t$. We also define $N_{\rm test}$ as the total number of test samples from the $T$ tasks.

\subsection{Synthetic Image Datasets for {\em Premonition}}\label{SM:synthetic_datasets}

The three primary {\em realm} datasets we created using Stable Diffusion are summarized in~\cref{T_datasets}.
\begin{table}[h]
\caption{{\bf Summary of synthetic {\em realm} datasets.}  These datasets were used to obtain {\em Premonition} results tabulated in~\cref{T_primary}. The column headings are: $n$ is the number of samples per class, $K$ is the number of classes and $N$ is the total number of samples.}\label{T_datasets}
\centering
\begin{tabular}{l|ccc}
 \toprule
{\bf {\em realm}} & $n$ & $K$ & $N$ \\
\toprule
`birds' & 200& 620 & 124,000 \\
\midrule
`food' & 680& 310 &210,800\\
\midrule
`plants' & 200& 1010 & 202,000\\
 \bottomrule
\end{tabular}
\end{table}

\subsection{Real Image Datasets}\label{SM:real_datasets}

\begin{figure}[h]
  \centering
  \includegraphics[width=0.8\columnwidth]{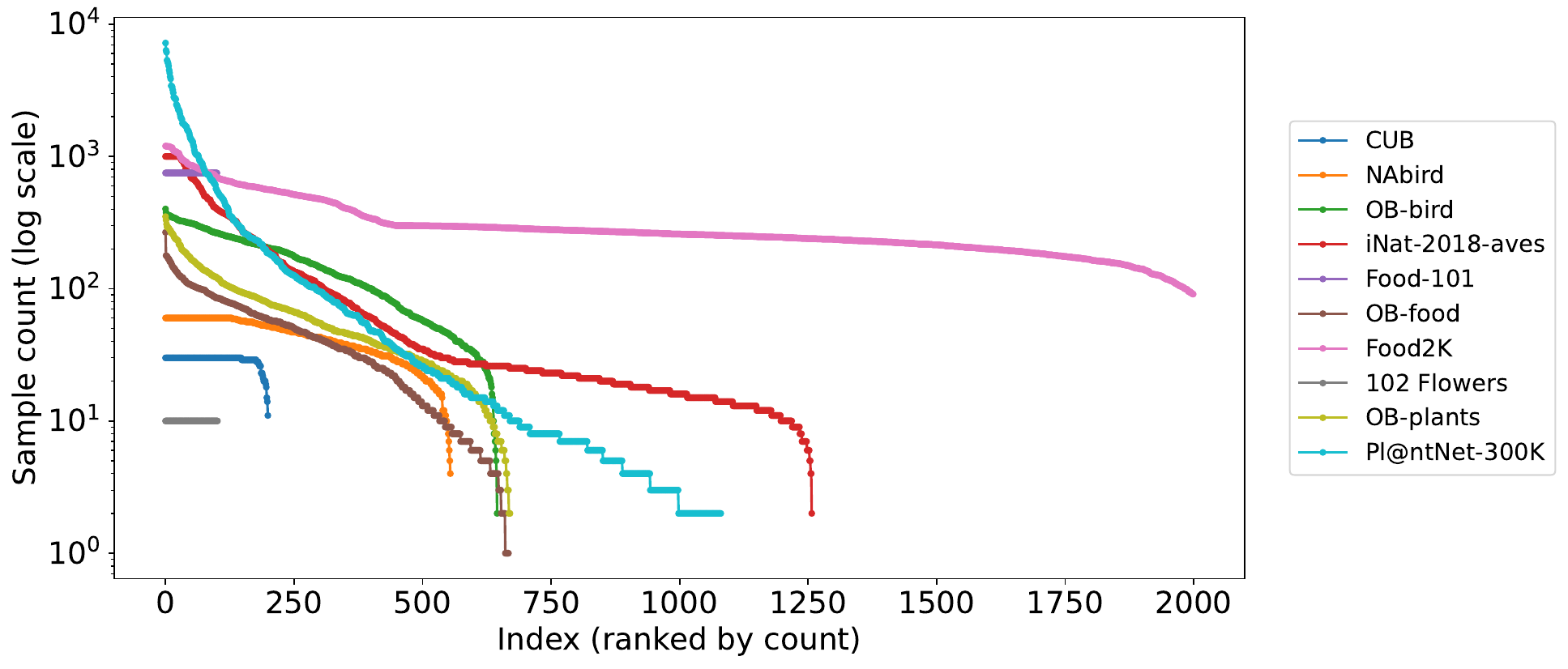}
  \caption{{\bf Distribution of class counts in training splits of datasets.} This figure illustrates that most of the datasets we demonstrate {\em Premonition} on have long tails where many classes have relatively few training samples. }\label{fig:long}
\end{figure}

The datasets\footnote{For NABirds~\cite{NAbird}, the following acknowledgement of use is required: ``{\em Data provided by the Cornell Lab of Ornithology, with thanks to photographers and contributors of crowdsourced data at AllAboutBirds.org/Labs. This material is based upon work supported by the National Science Foundation under Grant No. 1010818.}''} of real images used for CIL experiments tabulated in~\cref{T_primary} are shown in~\cref{T_datasets_CIL}. Different  to some related work~\cite{zhou2023revisiting,McDonnell23}, we use the standard CUB~\cite{CUB} train-test split.  \cref{fig:long} illustrates that most of the datasets are not only fine-grained, but also long-tailed, in that there are many classes with relatively few training samples. For example, Pl@ntNet-300K has more than 25\% of its classes with less than 10 training samples. 

\subsection{Synthetic Image Datasets for Data Replacement}\label{SM:synthetic_datasets_small}

For data replacement experiments, we created synthetic data for each class in the CUB, Food101 and 102Flowers datasets using Stable Diffusion, as per~\cref{T_datasets_small}. The number of samples per class is very small compared with synthetic data used for {\em Premonition} experiments (\cref{T_datasets}), to reflect the similar small number of training samples in the corresponding real datasets (see~\cref{T_datasets_CIL}).

\begin{table}[h]
\caption{{\bf Datasets used for CIL experiments.} As per~\cref{S:CIL_outline}, $K$ is the total number of classes, $N$ is the total number of training samples and $N_{\rm test}$ is the total number of test samples.}\label{T_datasets_CIL}
\centering
\begin{tabular}{c|cccc}
\toprule
{\bf Dataset}&  $N$& $N_{\rm test}$  &  $K$\\
\midrule
CUB~\cite{CUB}&$5794$ & $5994$& $200$ \\
NABirds~\cite{NAbird} &$23929$ &$24633$ & $555$\\
OB-bird~\cite{OmniBenchmark} & $97432$& $12834$& $646$\\
iNat-2018-aves~\cite{inat}& $143950$&$3774$ & $1258$\\
Food-101~\cite{bossard14}&$75750$ & 25250& $101$\\
OB-food~\cite{OmniBenchmark} & $30541$&$12728$ & $673$\\
Food2K~\cite{food2k} & $620192$& $104513$& $2000$\\
102Flowers~\cite{102flowers} & $1020$& $6149$& $102$\\
OB-plants~\cite{OmniBenchmark} &$45751$ & 13197& $671$\\
Pl@ntNet-300K~\cite{plantnet-300k} & $243916$&$31118$ & $1081$\\
\bottomrule
\end{tabular}
\end{table}

\begin{table}[h]
\caption{{\bf Summary of synthetic real datasets.}  These datasets were used to obtain Data Replacement results tabulated in~\cref{T_dr}. The column headings are: $n$ is the number of samples per class, $K$ is the number of classes and $N$ is the total number of samples.}\label{T_datasets_small}
\centering
\begin{tabular}{l|ccc}
 \toprule
{\bf {\em realm}} & $n$ & $K$ & $N$ \\
\toprule
CUB & 50& 200 & 10000 \\
\midrule
Food-101 & 50& 101 &5050\\
\midrule
102Flowers & 10& 102 & 1020\\
 \bottomrule
\end{tabular}
\end{table}

\clearpage
\section{CIL Methods}\label{S:Appendix_methods}

\subsection{GPT-4 Prompting}\label{SM:prompts}

Parsing the output of a chat/instruct style LLM can be challenging due to high variability in the LLM responses. To mitigate this, we used two strategies. First, we used `system' prompts to ask the LLM to provide structured outputs, in the form of lists or csv file text strings. Second, we used `few shot prompting', in the form of providing exemplars in the context passed to the LLM of a `user' prompt and `assistant' response. Further details of these strategies can be found in the public repository for this paper at \url{https://github.com/cl-premonition/premonition}.\\
~\\
{\bf Initial `system' prompt for subtypes of a Realm:} {\em Premonition} commences with a single short description for its input, for the name of a {\em realm}, such as `Birds', `Foods', `Plants', `Items found near roads', `Domestic furniture', or `Insects.'  The initial step of {\em Premonition} is to generate a list of subtypes within the realm, each of which then is used to prompt for fine-grained class names and descriptions. The initial prompt used for the realm `Birds' was:
\begin{quote}
List the latin names of all scientific orders of Birds, ensuring every item in the list is unique. Do not use extra explanations like `Please note that this is not an exhaustive list.' If the provided scientific order was already asked of you, simply repeat your previous answer for it.
\end{quote}
For non-biological realms, we use a slightly different initial prompt, e.g. for the {\em realm} `Items found near roads', or the {\em realm} `Food' we would use:
\begin{quote}
For each provided prompt, provide a list of 20 of the most common subcategories, ensuring every item in the list is unique. Do not use extra explanations like `Please note that this is not an exhaustive list.' If the provided prompt was already asked of you, simply repeat your previous answer for it.
\end{quote}
{\bf Description `system' prompt:} The GPT-4 `system' prompt we used for generating classes and descriptions for scientific orders of the `birds' {\em realm} is:
\begin{quote}
I want you to act as an input text prompt for the generative image model called Stable Diffusion. I will give you the scientific name of an order of Birds. Your job is to provide a numbered list of 20 unique randomly chosen *non-extinct* species from the provided order, and for each species give a detailed descriptions in 180 characters or less of the visual characteristics that will help to tell it from other species in a photograph. Also provide each species' latin scientific name, and if known, the common name. Provide them in csv format using ; as the separator with the following columns: item\_id, species\_latin\_name, species\_common\_name, species\_description. Do not use ; within any text fields. If the order has less than 20 species, list all the species using a shorter list. Do not list extinct species or any species from orders different to the one I provide. Do not explain if the order has less than 20 species. Ensure every listed species is only listed once. Do not explain using a note of the form `Note: The X order only contains these Y species.' Do not use ** in the result. If the same species was already asked of you, simply repeat your previous answer for it. Describe the visual characteristics and leave out how they sound or smell.
\end{quote}
This prompt was used in multiple repeats, with different bird order names provided as the `user' prompt. For realms that were not `biological', we used slightly different system prompts. The complete GPT-4 `system' prompt we used for the `Food' {\em realm} is as follows:
\begin{quote}
I want you to act as an input text prompt for the generative image model called Stable Diffusion. I will give you the name of a physical object/concept related to the theme of Food. Your job is to provide a numbered list of 20 unique randomly chosen types/kinds/examples of the object/concept, and for each species give a detailed descriptions in 180 characters or less of the visual characteristics that will help to tell it from other related items in a photograph. Provide them in csv format using ; as the separator with the following columns: item\_id, item\_name, item\_description. Do not use ; within any text fields. If the object/concept has less than 20 unique types/kinds/examples, list all of them using a shorter list. Do not list items unrelated to the object/concept to the one I provide. Do not explain if the object/concept has less than 20 type/kind/example. Ensure every listed item is only listed once. Do not use ** in the result. If the same item was asked of you, simply repeat your previous answer for it. Describe the visual characteristics and leave out how they sound, smell, taste or feel.
\end{quote}
~\\
{\bf Class `system' prompt:} Where we assume the class names are known prior to prompting GPT-4 for descriptions, we use a  prompt the commences with the following form:
\begin{quote}
I want you to act as an input text prompt for the generative image model called Stable Diffusion. I will give you a list of the English name of ten bird species. Your job is to provide a detailed descriptions in 250 characters or less of the visual characteristics of each species that will help to tell it from other species in a photograph....
\end{quote}
For example, for the CUB class names, we used this prompt (plus further formatting instructions like those above) twenty times with a different list of ten birds in the `user' prompt each time.

\subsection{Pre-training on Synthetic Data}\label{SM:PT}

For ResNet50 models with existing pre-trained weights we fine-tuned on our synthetic data by updating all parameters in the entire model using SGD and crossentropy loss with softmax. We used a  constant learning rates of $0.01$ in the output head and $0.0001$ in the backbone, a batch size of $128$ and momentum of $0.9$. The number of epochs used varied for different datasets. Models  trained from scratch instead used an initial learning rate of $0.05$ for all parameters, weight decay of $0.0005$, and a cosine annealing schedule that finished with a learning rate of $0$. Generally we trained for $60$ epochs.

\subsection{CIL Methods}\label{SM:CIL}

In the Nearest Class Mean (NCM) method, Class Prototypes (CPs) after task $T$ are a {\em mean} feature vector, namely:
\begin{equation}\label{CP}
    {\bf \bar{c}}_y=\frac{1}{n_y}\sum_{t=1}^T\sum_{i=1}^{N_t}{\bf h}_{t,i}\mathcal{I}_{t,i}\quad y=1,\dots,K,
\end{equation}
where  $\mathcal{I}_{t,i}$ is an indicator function with value $1$ if the $i$--th training sample in the $t$--th task is in class $y$ and zero otherwise,  $n_y$ is the number of training samples in each class and ${\bf h}_{t,i}$ is an $M$-dimensional projected and activated feature vector.  Classification during inference requires calculation of the cosine similarity between a sample's feature vector, and all class prototypes, 
\begin{equation}\label{cosine}
s_y=\frac{{\bf h}_{\rm test}^\top{\bf \bar{c}}_y}{||{\bf h}_{\rm test}||\cdot||{\bf \bar{c}}_y||},\quad y=1,\dots,K,
\end{equation}
and then selecting the $y$ with highest $s_y$. For continual LDA, We use the approach described in~\cite{panos2023session}.

The RanPAC method~\cite{McDonnell23} has two phases. The first involves first-session adaptation, but as described in the main text, we assume this is not viable. We therefore use only RanPAC's Phase 2. Except for class imbalance correction described below, we follow exactly the approach in~\cite{McDonnell23}: we use a frozen random projection layer of size $M=10000$ that increases the dimensionality of representations used for class prototypes from the backbone model's feature representation ($L=2048$ for ResNet50) to $M=10000$. The result is then nonlinearly activated with a ReLU function and its outer-product then calculated. The Gram matrix, ${\bf G}$, which is updated interatively one training sample at a time, can be written as ${\bf G}={\bf H}{\bf H}^\top$, where ${\bf H}$ is a matrix formed from the concatenation of all ${\bf h}$ from each training sample passed through RanPAC's random projection layers. We also have a matrix  ${\bf C}$ formed from the concatenation of all class prototypes defined as sums of feature vectors for each class. At the end of each task, the linear output classification head is  computed as ${\bf W}_{\rm o}=({\bf H}{\bf H}^\top+\lambda{\bf I})^{-1}{\bf C}$, where $\lambda$ is a regularization parameter calculated as described in~\cite{McDonnell23}.

\subsubsection{RanPAC for Imbalanced Class Distributions}

In this paper we have considered more challenging datasets than used in comparison work~\cite{zhou2023revisiting,McDonnell23,zhang2023slca,panos2023session}, \ie we consider only fine-grained datasets, some of which have severe class imbalance (see~\cref{fig:long}). When calculating class prototypes using the approach of~\cite{McDonnell23}, we found it was important for best results to modify that approach to account for class imbalance, as follows. We first define the {\em class frequency} for class $y$ as $\pi_y$, $y=1,\dots K$, \ie $\pi_y$ is the total number of samples in class $y$ observed up to the current point in time during continual learning.  We then define a $K \times K$ diagonal matrix ${\bf V}$ such that  element ${\bf V}_{k,k}$ is the reciprocal of the class frequency, \ie~${\bf V}_{k,k}=1/\pi_y$. Each class prototype becomes identical to $\bar{{\bf c}}_y$ used in NCM (i.e. means rather than  sums), while the decorrelated class prototypes becomes
\begin{eqnarray}\label{Imb}
    {\bf W}_{\rm o}
    &=&({\bf H}{\bf V}{\bf H}^\top+\lambda{\bf I})^{-1}\bar{{\bf C}},
\end{eqnarray}
where $\bar{{\bf C}}$ is the concatenation of the $\bar{{\bf c}}_y$ for all classes.

\section{Supplementary Results}\label{S:Appendix_results}

\subsection{Split CIFAR-100 and Split ImageNet-R}\label{S:split}

\cref{T_CIFAR} shows data for experiments with  commonly used CIL benchmarks, split CIFAR100~\cite{L2P,zhou2023revisiting}  and split ImageNet-R datasets~\cite{DualPrompt}, for $T=10$ tasks and four pre-trained models, two of which are trained only on synthetic data from Stable Diffusion. We apply {\em Premonition}'s fine-tuning step and Phase 2 of RanPAC~\cite{McDonnell23} with $M=10000$. For split-CIFAR100, the Stable Diffusion pre-trained model's final average accuracy falls short of the performance of models pre-trained on real data.   However, for ImageNet-R, the  model pre-trained only on synthetic ImageNet data~\cite{Sariyildiz_2023_CVPR} performs best.This preliminary investigation provided us with early evidence that pre-training on synthetic data can be beneficial for CIL, but we note that the pre-trained model of~\cite{Sariyildiz_2023_CVPR} required generation and training on many more samples than the real ImageNet, reflecting the diversity of concepts in the ImageNet dataset.

\begin{table}[b]
\caption{{\bf CIL results using {\em Premonition} fine-tuning with RanPAC~\cite{McDonnell23} for various ResNet50 pre-trained models.} Data shown is  final Average Accuracies~\cite{NIPS2017_f8752278} for $T=10$ CIL tasks. The divider after LGSSL separates pre-training on real imagery from pre-training on synthetic imagery. SD indicates pre-training was only on synthetic imagery from Stable Diffusion.}\label{T_CIFAR}
\centering
\begin{tabular}{l|c|cc}
 \toprule
{\bf Pre-training} & {\bf self-sup.} & {\bf CIFAR100} & {\bf IN-R} \\
\midrule
 IN1K (real) & &77.5 &  56.8\\
LGSSL~\cite{El_Banani_2023_CVPR} (real) &\checkmark & 67.2& 47.3\\
\midrule
IN1K-SD~\cite{Sariyildiz_2023_CVPR} & &61.3& {\bf 56.9} \\
CIFAR100-SD~\cite{shipard2023diversity} & &64.3 & 19.0\\
 \bottomrule
\end{tabular}
\end{table}

\begin{table}[t]
\caption{{\bf Comparison of pre-trained models for RanPAC.} The baseline model is the IN1K pre-trained ResNet50. The comparison models are {\em Realm} models seeded with the same IN1K pre-trained ResNet50 and then fine-tuned on synthetic data from the `birds', `food', and 'plants' {\em realms}.  }\label{T_cross}
\centering
\begin{tabular}{c|c|c|c|c|c|c}
 \toprule
 {\bf Pre-trained model} & {\bf CUB} & {\bf Food-101} & {\bf 102Flowers} &{\bf iNat-Aves} &{\bf Food2K} & {\bf Pl@ntNet}\\
\midrule
Baseline (IN1K) & $68.3$& $69.4$& $83.2$ & $37.4$&$46.4$ &$37.6$\\
\midrule
IN1K+`birds' & ${\bf 74.7}$& $65.3$&$83.0$  & {\bf 42.2}&$45.5$ &$37.9$\\
IN1K+`food' & $63.1$ &${\bf 71.7}$ & $86.8$ &$27.0$ & ${\bf 53.7}$&$40.0$\\
IN1K+`plants' & $66.6$ &$67.6$ &${\bf 90.2}$  &$31.3$ & $48.6$&${\bf 44.3}$\\
 \bottomrule
\end{tabular}
\end{table}

\subsection{Realm-Specific and Multi-Realm Pre-training}\label{SM:cross}

In order to check that the synthetic imagery used in {\em Premonition} needs to be tailored to a specific {\em realm}, we  investigated the case of a {\em Premonition} model created for one {\em realm} applied to another. The resulting data in~\cref{T_cross} for six of the datasets (the smallest and largest in each {\em Realm}), RanPAC, and an initial model seeded with IN1K pretrained weights, shows that surpassing the performance of the IN1K pre-trained model clearly requires {\em realm}-specific synthetic imagery. Indeed, when models from a mis-matched {\em realm} are used, the performance mostly goes backwards relative to not using {\em Premonition} at all, while always being significantly worse than using the desired {\em realm}. We conclude that it is important and of value to design a synthetic dataset relevant to the {\em realm} anticipated for a continual learner.

The converse of the above is the case where {\em Premonition} uses data from multiple unrelated {\em realms} and the resulting model applied to {\em realm}-specific datasets. To investigate this, \cref{T_multi} shows comparisons between the strongest results of~\cref{T_primary}, with data for the case where {\em Premonition} made use of data from a multi-theme {\em realm}, \ie~where synthetic data used for pre-training included data for the {\em realm}: `birds, food and plants.' We used RanPAC~\cite{McDonnell23} and  the entirety of our three synthetic datasets for {\em Premonition}, meaning the model was pre-trained on 1940 classes and over 500K samples. It can be seen that pre-training on this extra data for models initialized using IN1K actually enhances the value of {\em Premonition} for the Food and Plants datasets. For  Birds, the results are slightly worse than pre-training only on Birds, but still surpasses the Baseline. Possibly this is because Birds forms only 20\% of the 3-{\em realm} synthetic data, whereas Food and Plants are each 40\%. The use of multi-{\em realm} pre-training also enhances the value of {\em Premonition} on all but one dataset (Food-101) when training from scratch. 

\subsection{Zero-shot CLIP Classification for Fine-Grained Datasets}\label{SM:CLIP_ZS}

\subsubsection{Descriptions Versus Only Class-Names}

Stable Diffusion used in its default text-to-image mode uses a CLIP language model to encode text prompts. Hence, an obvious alternative to using Stable Diffusion to generate training data for {\em Premonition} is to combine its language model with the corresponding CLIP vision model, to do zero-shot classification, using prompts corresponding to class names, and to assess performance on the downstream test set. To create this baseline, we use the GPT-4 model to generate both Class (C) and Description (D) type text prompts (\cref{SM:prompts}) for each class in each of our target fine-grained datasets.  
As per the main text, D prompts are of the form: {\tt A photograph of a *class name*: *class description*} and C prompts of the form {\tt A photograph of a type of *{\em realm} name*: *class name*}. We use the standard CLIP zero-shot method~\cite{Radford} and tabulate zero-shot accuracies in~\cref{tab:Tab_zero_shot1}. In some cases, accuracy increases significantly for D prompts. The biggest improvement is in iNaturalist-2018-aves for `birds'. This can be readily traced to the fact that with this dataset, C prompts include only the latin scientific name of a species, resulting in the CLIP model's accuracy being very low compared to when the English common name is used within the description. In four cases (NABird, Food101, OB-food and OB-plants), including a description actually causes worse performance.

\begin{table}[tb]

\caption{{\bf {\em Premonition} results for pre-training on multi-theme realms.} Data is for the RanPAC CIL method~\cite{McDonnell23}. The first three rows are for models intialized with IN1K pre-trained weights. The final two rows were models trained on synthetic data in {\em Premonition} from scratch. `Single {\em realm}' means {\em Premonition} was applied such that the synthetic data used was generated for the same realm as the CIL dataset, \eg~CUB was tested using a model primed on only synthetic `Birds' data. `Multi {\em realm}' means that premonition was applied using synthetic data from all three realms.
}\label{T_multi}
\centering

\begin{tabular}{p{2.4cm}>{\centering\arraybackslash}p{1.5cm}|p{0.7cm}p{0.7cm}p{0.7cm}p{0.7cm}|p{0.7cm}p{0.7cm}p{0.7cm}|p{0.7cm}p{0.7cm}p{0.7cm}}

&&\multicolumn{4}{|c|}{\bf Birds {\em realm}}& \multicolumn{3}{|c|}{\bf Food {\em realm}}& \multicolumn{3}{|c}{\bf Plants {\em realm}}\\
&
  & \STAB{\rotatebox[origin=c]{90}{CUB}} & \STAB{\rotatebox[origin=c]{90}{NABird}} & \STAB{\rotatebox[origin=c]{90}{OB-bird}} & \STAB{\rotatebox[origin=c]{90}{iNat-2018-aves}} & \STAB{\rotatebox[origin=c]{90}{Food-101}} & \STAB{\rotatebox[origin=c]{90}{OB-food}} & \STAB{\rotatebox[origin=c]{90}{Food2K}} & \STAB{\rotatebox[origin=c]{90}{~102 Flowers}} & \STAB{\rotatebox[origin=c]{90}{OB-plants}} & \STAB{\rotatebox[origin=c]{90}{~Pl@ntNet-300K}} \\
  \bottomrule
  \end{tabular}
\vspace{-2mm}
\begin{tabular}{p{3.8cm}>
{\centering\arraybackslash}p{0.1cm}|p{0.7cm}p{0.7cm}p{0.7cm}p{0.7cm}|p{0.7cm}p{0.7cm}p{0.7cm}|p{0.7cm}p{0.7cm}p{0.7cm}}
Baseline (No {\em Premonition)}  && $68.3$&$57.0$ & $44.8$&$37.4$ &$69.4$ &$19.5$ &$46.4$ & $83.2$& $33.0$&$37.6$  \\
\midrule
IN1K + single {\em realm}  && ${\bf 74.7}$& ${\bf 64.9}$& ${\bf 49.4}$& ${\bf 42.2}$&${ 71.7}$ & ${ 19.5}$&${ 53.7}$ & ${ 90.2}$& ${ 36.7}$& ${ 44.3}$ \\
IN1K + multi {\em realm}&&${74.3}$ & $64.6$&$49.2$ &$39.4$ & ${\bf 72.0}$& ${\bf 20.3}$& ${\bf 53.7}$&${\bf 91.1}$ & ${\bf 38.1}$& ${\bf 45.1}$\\
\midrule
Scratch + single {\em realm} && $60.8$& $49.1$& $39.0$&$22.2$  &$70.3$ &$16.0$ & $48.6$& ${87.4}$& ${35.1}$& ${39.0}$\\
Scratch + multi {\em realm} && $64.4$&$49.5$ & $39.2$& $26.3$& $68.7$& $17.7$&$49.7$ &$89.9$ & $37.1$& $41.7$\\
\bottomrule
%\vspace{3pt}
\end{tabular}
\end{table}

\begin{table}[tb]
    \caption{{\bf CLIP ViT-L/14 zero-shot baselines.} Data is zero-shot accuracy on the real datasets' test splits.  C prompts mean class names were used within the CLIP text prompt; D prompts mean descriptions generated by GPT-4 were used within the CLIP text prompt. As this is zero-shot classification. we use only the test/validation set for each dataset.
    }
    \label{tab:Tab_zero_shot1}
    \centering
    \begin{tabular}{c|c|c|c}%|c|c}
    \toprule
    {\bf {\em Realm}} & {\bf Dataset}  & {\bf Prompt}&{\bf Acc.} \\
    \midrule
        `birds' & CUB &  C& 62.6 \\
        & & D& 65.0 \\
        &NABird &   C& 51.4 \\
         & & D&50.6\\
        &OB-bird &  C& 33.8\\
         & & D& 38.8\\
        &iNat-2018-aves&  C& 7.8 \\
         & & D&33.2\\
       \midrule
        `food' & Food101 &   C&93.5 \\
         & & D&88.6\\
       & OB-food  &  C&38.5 \\
        & & D&35.7\\
        & Food2K &  C&11.3 \\\
        & & D&11.9\\
        \midrule
        `plants' & 102 Flowers  &   C&77.2\\
         & & D&85.6\\
        & OB-plants  & C& 39.2\\
         & & D&36.4\\
        & Pl@ntNet  &  C&15.4\\
         & & D&17.7\\
        \bottomrule
    \end{tabular}
\end{table}

\subsubsection{Zero-Shot Versus {\em Premonition}}

Only two datasets in~\cref{tab:Tab_zero_shot1}  have CLIP zero-shot accuracies higher than the best {\em Premonition} results in~\cref{T_primary}. Although these are both for `food' datsets (Food101 and OB-food), for Food2K, the zero-shot performance is very low compared to all other results. Noting that Food2K has 2000 classes, the latter result is consistent with remarks in~\cite{Radford} on the difficulty that CLIP zero-shot classification has with fine-grained classification tasks. For all the `birds' and `plants' datasets, CLIP zero-shot performance is surpassed by the main {\em Premonition} results shown in~\cref{T_primary}.  This is despite the CLIP model being a much large transformer network model than the ResNet50 model we use.

\subsubsection{CLIP Zero-Shot Accuracy on Synthetic Images}\label{SM:CLIP_synthetic}

One measure of the quality of our synthetic imagery is to assess the performance of CLIP zero-shot classification on these images using the same prompts used to generate them. The results in~\cref{T_CLIP}, for the dataset in each {\em realm} for which zero-shot accuracy on the real dataset is strongest (and with the fewest classes), show there is a significant drop in raw zero-shot accuracy for the synthetic data compared to the real data (using D prompts). This indicates there is potential value in seeking to improve the quality and discriminability of generated synthetic images used with {\em Premonition}. 

\begin{table}[tb]
\caption{{\bf CLIP zero-shot accuracy on Stable Diffusion generated datasets.} The {\bf Drop} column quantifies the substantial drop in accuracy for CLIP zero-shot classification on synthetic imagery compared with the corresponding real validation set zero-shot accuracies (see~\cref{tab:Tab_zero_shot1}). We used the same text prompt and CLIP VIT-L/16 language model for CLIP zero-shot classification as used to generate the synthetic data from Stable Diffusion. }\label{T_CLIP}
\centering
\begin{tabular}{c|c|c}
 \toprule
{\bf Dataset}  &{\bf Accuracy (\%)} & {\bf Drop (\%)}\\
\midrule
CUB & $47.8$ & $-17.2$\\
\midrule
Food-101  & $76.2$& $-12.4$\\
\midrule
102 Flowers &$60.7$ &$-24.9$ \\
 \bottomrule
\end{tabular}
\end{table}

\subsection{Discriminating Synthetic and Real Images}\label{SM:binary}

The Synthetic Gap~\cite{Zhang_synthetic_gap} discussed in the Introduction is consistent with a body of work that has studied the ease with which synthetic data can be discriminated from real data by trained binary classifiers. For example,  Bird and Lotfi \cite{bird2023cifake} created a synthetic version of CIFAR-100 using Stable Diffusion, and trained classifiers that could distinguish real from synthetic with high accuracy. They analyse why, and conclude that ``the model focuses on small visual imperfections in the background of the images.''  In their literature review, they discuss prior work that found that ``images generated by various latent diffusion approaches may contain digital fingerprints to suggest they are synthetic''~\cite{sha2023defake}. The latter paper analysed multiple generative models, and concluded that artifacts created by each not only enable detection of synthetic imagery, but also attribution to a particular model.  Consistent with this line of work, the results  in~\cref{T_binary} show that the synthetic data we generate using Stable Diffusion is easy to discriminate from real data. This is despite the semantic content of the test sets  not being exposed to the model during training.  The ease with which synthetic data can be discriminated from real data helps explain why we found that the use of synthetic data for data replacement was not able to assist continual learning.

To produce~\cref{T_binary}, we created three binary datasets and trained and tested binary classifiers for each. Each dataset had $40000$ training samples and $20000$ test samples. The training sets had $20000$ real images and $20000$ synthetic images; the test sets had $10000$ of each.  In order to make the model's task as challenging as possible within the context of our main use of synthetic data, we assigned two out of the three {\em realms} to the training set and the third to the test set. For example, the first dataset included synthetic `birds' and `plants' images in the training set, and real images selected from the 4 real `birds' datasets and the 3 real `plants' dataset. The corresponding test set used synthetic `food' images and real `food' images from the three real food datasets. All images for each dataset were selected  uniformly randomly from each class in each dataset used, until the desired number of samples was achieved.  The model we trained was a ResNet50 initialized with IN1K pre-trained weights. We used a learning rate of $0.05$, weight decay of $0.0005$, momentum of $0.9$, a cosine learning rate schedule with final learning rate of $0$ and trained for $5$ epochs.

\clearpage
\begin{table}[tb]
\caption{{\bf Binary classification of synthetic/real images.} These results show that synthetic imagery is discriminable from real imagery with high accuracy, even when the model is trained on data from a different {\em realm} to that which it is tested on. }\label{T_binary}
\centering
\begin{tabular}{c|c|c}
 \toprule
{\bf Training {\em realms}}  & {\bf Test {\em realm}} & {\bf Accuracy (\%)}\\
\midrule
`birds' \& `plants' & `food' & 99.2\\
`birds' \& `food' & `plants' & 93.1\\
`food' \& `plants' & `birds' & 97.4\\
 \bottomrule
\end{tabular}
\end{table}

\subsection{Comparison With Using Only Class Names}\label{SM:class_name}

\cref{T_vs} shows data referred to in~\cref{S:Discussion}, where we assume, unlike the main text, that {\em Premonition} can know all class names in advance. The three synthetic datasets achieved better performance than the baseline {\em Premonition} results.

Also shown in~\cref{T_vs} is data analogous to the Data Replacement experiments of~\cref{S:DR}. Since in this section the same class names are used both in pre-training on synthetic data, and the downstream CIL dataset, we used the classification head of the pre-trained model to classify the test set of the corresponding real datasets. Note that this does not use any training data from the real datasets, and classification is therefore zero-shot. It can be seen that the accuracies are very poor, yet again highlighting that synthetic imagery has a substantial domain gap.

\vspace{-0.5cm}
\begin{table}[h]
\caption{{\bf RanPAC CIL for $T=10$ tasks using synthetic data with prior knowledge of classes.} In the table header, {\bf Baseline} accuracies are less than shown for {\em Premonition} in ~\cref{T_primary}, because fewer training samples were used here, to match the class-name synthetic datasets. {\bf Class Premonition} means comparison models trained on  synthetic data with the same classes as the real dataset and  {\bf Data Replacement} means results when the output head from training on synthetic data was used to classify the real data in the corresponding dataset, without any new learning on the real data, \ie zero-shot non-continual. Models were ResNet50s initialized with IN1K weights.}\label{T_vs}
\centering
\begin{tabular}{c|c|c|c|c|c|c}
 \toprule
{\bf Dataset}  & $n$& $K$& $N$& {\bf Baseline (\%)} & {\bf  Class Prem. (\%)} & {\bf Data rep. (\%)}\\
\midrule
CUB & $640$&$200$ &$128,000$ &$74.7$ &$78.0$ & $29.5$\\
Food-101&$640$ &$101$ &$64640$  &$68.6$ & $69.1$& $38.8$\\
102 Flowers& $1000$& $102$& $102000$ &$88.0$ & $89.7$& $46.6$\\
 \bottomrule
\end{tabular}
\end{table}
\vspace{-0.5cm}

\subsection{ViT Pre-trained Models}\label{SM:VIT}

\cref{T_primary_extra}  shows results for pre-trained transformer networks, namely ViT-B/16. For {\em Premonition} with ViT, we use the Adaptformer~\cite{chen2022adaptformer} Parameter-Efficient Transfer Learning (PETL) method trained purely on synthetic data from the {\em realm} corresponding to each dataset. For RanPAC, {\em Premonition} always provides a gain, albeit relatively small compared with those for ResNet50 in~\cref{T_primary}.  Notably, all `birds' and `plants' ViT-B/16 models surpass the CLIP zero-shot baseline by large margins. For the `food' domain, Food2K results all surpass the baseline. However no models do so for OB-food or Food-101. 

\vspace{-0.5cm}
\begin{table}[h]
\caption{{\bf {\em Premonition} results for three CIL methods applied to 10 datasets.} Results for ViT-B/16, pre-trained on ImageNet-21K (IN21K) using self-supervision.  CLIP ViT-L/14 zero-shot performance for the same dataset sets is shown as a baseline.
}\label{T_primary_extra}
\centering
\begin{tabular}{p{2.2cm}>{\centering\arraybackslash}p{1.5cm}|p{0.7cm}p{0.7cm}p{0.7cm}p{0.7cm}|p{0.7cm}p{0.7cm}p{0.7cm}|p{0.7cm}p{0.7cm}p{0.7cm}}

&&\multicolumn{4}{|c|}{\bf Birds {\em realm}}& \multicolumn{3}{|c|}{\bf Food {\em realm}}& \multicolumn{3}{|c}{\bf Plants {\em realm}}\\
&
  & \STAB{\rotatebox[origin=c]{90}{CUB}} & \STAB{\rotatebox[origin=c]{90}{NABird}} & \STAB{\rotatebox[origin=c]{90}{OB-bird}} & \STAB{\rotatebox[origin=c]{90}{iNat-2018-aves}} & \STAB{\rotatebox[origin=c]{90}{Food-101}} & \STAB{\rotatebox[origin=c]{90}{OB-food}} & \STAB{\rotatebox[origin=c]{90}{Food2K}} & \STAB{\rotatebox[origin=c]{90}{~102 Flowers}} & \STAB{\rotatebox[origin=c]{90}{OB-plants}} & \STAB{\rotatebox[origin=c]{90}{~Pl@ntNet-300K}} \\
  \end{tabular}
\begin{tabular}{p{2.4cm}>{\centering\arraybackslash}p{1.3cm}|p{0.7cm}p{0.7cm}p{0.7cm}p{0.7cm}|p{0.7cm}p{0.7cm}p{0.7cm}|p{0.7cm}p{0.7cm}p{0.7cm}}
 \midrule
 CLIP zero-Shot &-& $65.0$ &$51.4$ & $38.8$ & $33.2$ & $93.6$ & $38.5$ & $11.9$& $85.6$ & $39.2$ & $17.7$\\%& $13.3$\\
\midrule
NCM &IN21K&$86.2$ &$79.0$ & $55.3$&$56.3$ &$83.8$ &$30.9$ &$59.9$ & $99.4$& $56.4$&$55.7$ \\
\ditto~~+ {\bf Prem.}&IN21K&${\bf 86.9}$ & ${\bf 81.2}$& $55.1$& ${\bf 58.5}$& $82.8$&$30.3$ &58.1 &$99.3$ &$54.4$ & ${\bf 56.2}$\\
\midrule
Continual LDA&IN21K& $86.5$&$76.0$ & $54.4$& $53.1$&$84.4$ & $28.8$& $53.5$& $99.5$ &$57.1$ &$45.9$ \\
\ditto~~+ {\bf Prem.} &IN21K& ${\bf 86.9}$& ${\bf 77.2}$& ${\bf 54.5}$& ${\bf 53.9}$& ${\bf 84.9}$& ${\bf 29.0}$& ${\bf 53.8}$ &  ${\bf 99.6}$& $56.3$&$45.2$ \\
\midrule
   RanPAC &IN21K& $88.7$&$83.5$ & $63.0$& $63.5$&$89.1$ & $34.3$&$66.2$ & $99.5$ &$61.8$ &$64.0$ \\
\ditto~~+ {\bf Prem.}  &IN21K& ${\bf 89.8}$&${\bf 85.3}$ & ${\bf 64.4}$& ${\bf 65.8}$& ${\bf 89.4}$ & ${\bf 35.0}$& {\bf 66.9} & ${\bf 99.5}$ & ${\bf 62.1}$& ${\bf 64.6}$\\
\bottomrule
\end{tabular}
\end{table}
\vspace{-1cm}

\subsection{Other CIL Methods}\label{SM:CIL_extra}

{\em Premonition} can potentially enhance any existing CIL method where pre-trained models are used. Our choice of CIL methods (NCM, LDA and RanPAC) was based on three factors: (i) these three methods are specifically designed for leveraging strong pre-trained models (ii) each method is simple to implement and (iii) no special methods are needed for combating forgetting. However, as shown in~\cref{T_EWC}, it is not difficult to demonstrate the benefits of {\em Premonition} for other CIL methods. Here, for example, we use SLCA finetuning~\cite{zhang2023slca} and EWC~\cite{EWC}, and compare each method with a plain IN1K pre-trained model and a {\em Premonition} pre-trained model initialised with IN1K weights. 
\vspace{-0.5cm}
\begin{table}[h]
\caption{{\bf {\em Premonition with other CIL methods}. } Indicative results for CUB and Food-101, which illustrate that {\em Premonition} is useful regardless of the CIL method. All methods commenced with IN1K pre-trained weights and ResNet50.}\label{T_EWC}
\centering
\begin{tabular}{l|c|c}
\toprule
{\bf Method }    & {\bf CUB} & {\bf Food-101} \\
     \midrule
 SLCA Finetuning   & $61.4\%$ & $68.5\%$\\
 SLCA Finetuning + {\bf Premonition}   &  ${\bf 64.0\%}$&${\bf 69.5\%}$ \\ 
 \midrule
 EWC & $39.7\%$&$65.4\%$ \\
  EWC + {\bf Premonition}& ${\bf 43.5\%}$&${\bf 66.2\%}$\\
  \bottomrule
\end{tabular}
\end{table}

\clearpage

\end{document}